\documentclass{svproc}
\usepackage[utf8]{inputenc}
\usepackage{graphics} 
\usepackage{epsfig} 
\usepackage{amsmath} 
\usepackage{mathtools}
\usepackage{amsfonts}
\usepackage{amssymb}  
\usepackage{multirow, hhline}
\usepackage{algorithm}
\usepackage{algpseudocode}
\usepackage{lipsum}
\usepackage{subfloat}

\usepackage[export]{adjustbox}
\usepackage[dvipsnames]{xcolor}
\usepackage{soul,color}
\usepackage{mathdots}
\usepackage{caption, subcaption}

\usepackage{textcomp}
\usepackage{gensymb}
\usepackage{lettrine}
\usepackage{comment}

\usepackage{cite}

\allowdisplaybreaks

\makeatletter
\let\NAT@parse\undefined
\makeatother

\DeclareMathOperator*{\argmax}{arg\,max}

\usepackage{hyperref} 
\hypersetup{
  colorlinks   = true, 
  urlcolor     = blue, 
  linkcolor    = blue, 
  citecolor   = green 
}

\usepackage{url}

\mainmatter              
\begin{document}

\begin{center}
    This paper has been accepted for publication in the \textit{5th Iberian Robotics Conference (ROBOT)}.
\end{center}
\thispagestyle{empty}
\bigskip

© 2022 the authors.

\clearpage
\setcounter{page}{1}
\title{ExplORB-SLAM: Active Visual SLAM Exploiting the Pose-graph Topology
}
\titlerunning{ExplORB-SLAM}  
%
\author{Julio A. Placed \and Juan J. G\'omez Rodr\'iguez \and Juan D. Tard\'os \and Jos\'e A. Castellanos
}
\authorrunning{J. A. Placed \textit{et al.}
} 
%
\tocauthor{Julio A. Placed, Juan J. G\'omez Rodr\'iguez, Juan D. Tard\'os and Jos\'e A. Castellanos
}
\institute{University of Zaragoza, Zaragoza, 50018, Spain,
\\
\email{\{jplaced,jjgomez,tardos,jacaste\}@unizar.es}
}

\maketitle              

\begin{abstract}
    Deploying autonomous robots capable of exploring unknown environments has long been a topic of great relevance to the robotics community. In this work, we take a further step in that direction by presenting an open-source active visual SLAM framework that leverages the accuracy of a state-of-the-art graph-SLAM system and takes advantage of the fast utility computation that exploiting the structure of the underlying pose-graph offers.
    Through careful estimation of \emph{a posteriori} weighted pose-graphs, $D$-optimal decision-making is achieved online with the objective of improving localization and mapping uncertainties as exploration occurs.

\end{abstract}
%

\section{Introduction} \label{S:1}

\lettrine{A}{ctive} simultaneous localization and mapping (SLAM) names the problem of autonomously deciding where a robot should move next in order to form the most accurate representation of the environment (i.e., including localization and mapping) possible. Thus, it refers to the problem of decision-making on control commands for a robot which is performing SLAM and that is capable of navigating in an unknown environment. See~\cite{placed22} and the references therein.

Traditional approaches to solve active SLAM born within the robotics community are numerous and vary in shape and form~\cite{makarenko02,valencia12,umari17,carrillo18}. Most of them share, however, the following division in three discernible stages~\cite{feder99,placed22}:
\begin{itemize}
    \item Goal identification: to reduce the (ideally infinite) number of possible destinations, it is common to define a finite and typically low-dimensional subset of boundaries between the known and unknown regions, i.e., frontiers.
    \item Utility computation: the expected usefulness of executing a candidate set of actions can be estimated studying the expected uncertainty in the two target random variables; either leveraging information theory (IT) or theory of optimal experimental design (TOED).
    \item Selection and execution of the optimal action. After estimating utility, it all comes down to selecting the most useful destination and traveling to it.
\end{itemize}

The interest of this paper particularly lies on TOED-based approaches, that directly quantify utility from the variance of the variables of interest, under the assumption of Gaussianity. Feder \textit{et al.}~\cite{feder99} were the first to use task-oriented metrics (i.e., \emph{optimality criteria}) for active SLAM and, since then, their popularity has steadily grown due to their strong mathematical basis and elegant formulation. Their complexity and high computation burden, though, make information theoretic approaches still prevail in the literature~\cite{valencia12,deng20}.

In these active SLAM approaches, the second stage represents the main bottleneck as it requires analyzing the spectrum of the Fisher information matrices (FIMs) expected for a candidate set of actions. Such matrices must comprise the uncertainties of the robot state (e.g., pose) and the environment representation (i.e., map). For landmark-based maps, utility can be obtained either by studying a full (high-dimensional) FIM that contains both robot poses and landmarks~\cite{feder99}, or by computing the uncertainty of the robot subject to the set of observations, i.e., marginalizing the landmarks' information~\cite{carlone17}. In contrast to the above, Wang and Englot~\cite{wang20} incorporate the robot's expected variance into the landmarks' uncertainty. On the other hand, for dense metric maps (e.g., occupancy grids), most works resort to IT metrics, e.g., the entropy~\cite{valencia12}. An exception is the work from Carrillo \textit{et al.}~\cite{carrillo18}, where a TOED-based metric of the uncertainty in the robot location is used to correct the entropy of the map. The opposite idea is employed in~\cite{placed21iros}, where the authors heuristically embed the impact of the map's entropy on the location uncertainty metric.

For the case of active graph-SLAM, significant reductions in the time required to execute the second stage were recently achieved leveraging the insight of utility being closely linked to the topology of the underlying \emph{pose-graph}~\cite{khosoussi14,khosoussi19}. Intuitively, the connectivity (i.e., the Laplacian spectrum) of the pose-graph can be regarded as a measure of how well the environment model is being formed. Later works~\cite{placed21, placed21iros} demonstrate the tight relationship between modern optimality criteria and connectivity indices, when the edges of the graph are weighted appropriately. Hence, the optimal actions to execute can be found by studying the topology of the expected weighted pose-graphs rather than analyzing the full posterior FIM. Successful applications that leverage the above include, for example, coverage tasks with uncertainty awareness~\cite{chen20} or 2D active SLAM~\cite{placed21}.


\subsection{Contributions}

The main contribution of this work is an active visual SLAM system that exploits the topology of the underlying pose-graph to identify $D$-optimal candidate locations over affordable time horizons.
From the accurate sparse map and trajectory estimation ORB-SLAM2~\cite{mur17} provides, we design a decision-making mechanism that switches between \emph{exploration} and \emph{exploitation} principles. To do so, we adapt the insight of discounting information in Shannon-R\'enyi entropy~\cite{carrillo18} to operate in the task space, and leverage the theoretical results on pose-graphs' optimality from~\cite{placed21}. The above, along the method to predict reobservations and their associated information, emerge as side contributions of our work.
The last contribution is the code release in the \href{https://github.com/JulioPlaced/ExplORB-SLAM}{ExplORB-SLAM repository}
for ROS Noetic, representing a complete framework to test this and other active SLAM approaches and aiming to pave the way for reproducible research in this field.

\subsection{Paper Structure}

The rest of the paper is organized as follows. Section~\ref{S:2} contains a brief introduction to visual graph-SLAM, modern optimality criteria, and their relationship with graph connectivity indices. In Section~\ref{S:3}, we present and thoroughly describe the proposed system. Section~\ref{S:4} shows experimental validation results. Finally, the manuscript concludes in Section~\ref{S:5}, where future work is also outlined.

\section{Background}\label{S:2}

\subsection{Visual Graph-SLAM}



Visual SLAM refers to the problem of reconstructing the environment and locating an agent using only images from a set of cameras as inputs. This modality is a hot topic in both industry and academia as cameras are cheap sensors that also provide useful information for other computer vision tasks like object detection.

ORB-SLAM2~\cite{mur17} is one of the state-of-the-art algorithms in V-SLAM and is the basic building block for this paper. It
leverages the use of ORB features inside an accurate bundle adjustment (BA) optimization framework that minimizes the reprojection error of the estimated landmarks in order to refine its 3D position and the pose of the agent. In order to constrain the execution time,
ORB-SLAM only performs BA at a lower rate than the video frequency using a subset of the images received, so called \emph{keyframes}, that carry high visual innovation with respect the rest of the map.

An interesting property of BA is its sparsity pattern. With a simple study of its structure and its Hessian layout, one can realize that it has a sparse block-structure  that can be also regarded as a graph that connects keyframes to the observed landmarks.
To enhance computation efficiency of the algorithm, the landmarks can be marginalized out using the Schur complement obtaining the reduced camera system, whose Hessian only
relates keyframes that have common observations.
The corresponding graph is also simplified, forming
the so called \emph{pose-graph} in which vertices only represent keyframes and edges the relative pose between pairs of connected keyframes. ORB-SLAM, however, only builds this simplified graph when it needs to correct a loop. To reduce the computation burden, it uses an even sparser version of the pose-graph, the \emph{essential graph}. The peculiarity of this graph resides in the fact that it only connects keyframes when they share a minimum amount of observations. Thus, keyframes with few landmarks in common are not connected, sparsifying the graph.

\subsection{On the Optimality of SLAM Pose-graphs}

During the second step in active SLAM, quantifying uncertainty is of utmost importance. On the basis of TOED, Kiefer~\cite{kiefer74} proves the existence of a family of scalar mappings $\|\cdot\|:\mathbb{R}^{\ell\times\ell}\mapsto\mathbb{R}$ dependent of just the parameter $p$ and which can be expressed in terms of the eigenvalues ($\lambda$) of the matrix to be quantified, e.g., covariance ($\boldsymbol{\Sigma}$) or FIM ($\boldsymbol{\Phi}$):
\begin{equation}
    \|\boldsymbol{\Sigma}\|_p \triangleq \left( \frac{1}{\ell}\text{trace}(\boldsymbol{\Sigma}^p) \right) ^ \frac{1}{p}
\end{equation}
%
Particularizing the above equation for the boundary values of $p$, modern \emph{optimality criteria} result. The most relevant metric among them is $D$-optimality ($p=0$), since it captures the global uncertainty~\cite{kiefer74}. On the downside, it also entails the highest computational complexity. Its modern formulation is as follows:
\begin{equation}
    D{\text -} opt\left(\boldsymbol{\Sigma}\right) \equiv \|\boldsymbol{\Sigma}\|_0 \triangleq \exp \left(\frac{1}{\ell} \sum_{k=1}^\ell \log(\lambda_k) \right) \label{eq:dopt}
\end{equation}

Let us now consider the specific (and somehow prevalent) case in which a graph-SLAM algorithm is used during active SLAM. These methods use a graph representation in which nodes encode the robot and landmark states, and edges encode the constraints between them (e.g., observations, odometry or loop closures). A flatter representation can be achieved by marginalizing the observations and thus only representing the robot states in nodes, i.e. a pose-graph $\boldsymbol{\mathcal{G}}$. In such case, the $i$-th vertex $\boldsymbol{v}_i\in\mathcal{V}$ will contain the robot pose (in the world frame) $\boldsymbol{T}_{wi}\in SE(n)$. On the other hand, an arbitrary edge $\boldsymbol{e}_j\in\mathcal{E}$ that connects the pair of vertices $\{\boldsymbol{v}_i, \boldsymbol{v}_k\}\in\mathcal{V}$ will encode the relative transformation between such pair, $\boldsymbol{T}_{ik}\in SE(n)$, and its associated uncertainty. Due to its higher sparsity, the latter is preferably represented by the FIM $\boldsymbol{\Phi}_j\in\mathbb{R}^{\ell\times\ell}$; with $\ell$ the degrees of freedom of the $n$-dimensional Euclidean space.

Decision-making in active SLAM can be reduced to evaluating optimality criteria over the expected FIM of the entire system at candidate goal locations, in order to compare them and later travel to the most informative one. This full FIM $\boldsymbol{Y} \in \mathbb{R}^{|\mathcal{V}|\ell\times |\mathcal{V}|\ell}$ must comprise the uncertainty of both the map and the robot state, as discussed in Section~\ref{S:1}, and is commonly built upon the relative FIMs  $\boldsymbol{\Phi}_j$. See~\cite{placed21} for further details on its structure. High dimensionality of $\boldsymbol{Y}$ makes the analysis of its spectrum computationally intensive and even intractable for online systems with large state and/or action spaces.
By exploiting the facts that (i) the sparsity pattern of $\boldsymbol{Y}$ conveys that of the Laplacian of the underlying pose-graph and (ii) their spectra are intimately linked under certain conditions, optimality criteria of $\boldsymbol{Y}$ can be approximated by that of the (weighted) Laplacian of the graph. For the case of $D$-optimality, the following relationship may be established for the general case of graph-SLAM formulated over the Lie group $SE(n)$ \cite{placed21}:
\begin{equation}
    D{\text -} opt(\boldsymbol{Y}) \approx D{\text -} opt(\boldsymbol{L}_\gamma) = \left( |\mathcal{V}|\ t(\boldsymbol{\mathcal{G}}_\gamma)\right)^{\frac{1}{|\mathcal{V}|}} \label{eq:dopt_graph}
\end{equation}
\noindent where $\boldsymbol{\mathcal{G}}_\gamma$ is the posterior pose-graph in which each edge is weighted with the scalar $\gamma_j=\|\boldsymbol{\Phi}_j\|_0$, $\boldsymbol{L}_\gamma$ its Laplacian matrix, $t(\boldsymbol{\mathcal{G}}_\gamma)$ its weighted number of spanning trees and $|\mathcal{V}|$ the dimension of the set $\mathcal{V}$ (i.e., the total number of vertices).

\section{Proposed Method}\label{S:3}

\begin{figure}[t!]
    \centering
    \includegraphics[width=1\linewidth]{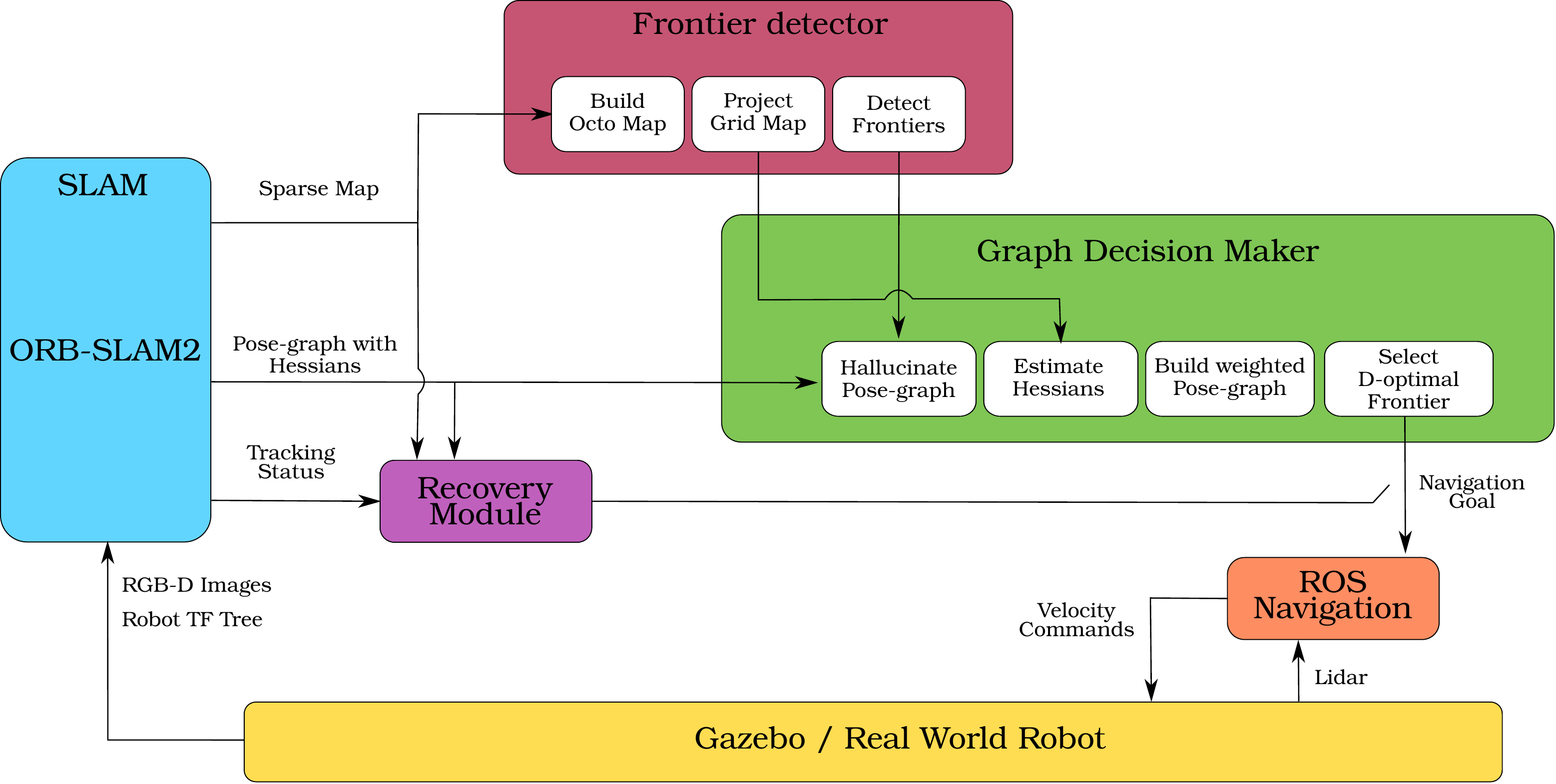}
    \caption{Overview of ExplORB-SLAM.}
    \label{fig:active_system}
\end{figure}

The proposed system leverages the fast computation of $D$-optimality using~\eqref{eq:dopt_graph} to achieve online autonomous exploration while performing visual graph-SLAM. It consists of several different modules, which are depicted in Figure~\ref{fig:active_system} and described hereafter. The whole system is built within a ROS Noetic framework that eases the communication between modules and allows the connection with Gazebo.

\subsection{SLAM}

This fundamental module contains a modified version of ORB-SLAM2 \cite{mur17} with a wrapper for ROS Noetic. In order to later perform decision-making, some changes were required. The most important one involves extracting the pose-graph and the information matrices\footnote{Throughout this paper, we will equally use FIM and Hessian matrix, since they are equivalent when evaluating the latter at the \textit{maximum likelihood estimate}.} associated to the relative movement between vertices (i.e., keyframes) so as to build the required weighted pose-graph.

In a separate thread, we construct the camera-point Hessian of the SLAM system using the Gauss-Newton approximation to the least squares problem, i.e., the Jacobians of the reprojection error as in a BA~\cite{triggs99}. It will have the form:
\begin{equation}
    \boldsymbol{H}_{\text{SLAM}}' =    \begin{pmatrix} \boldsymbol{H}_c & \boldsymbol{H}_{cp} \\
                                    \boldsymbol{H}_{cp}^T & \boldsymbol{H}_p
                    \end{pmatrix}
\end{equation}
\noindent where $\boldsymbol{H}_c$ and $\boldsymbol{H}_p$ are the blocks which define the information about the robot poses and the map points, respectively; and $\boldsymbol{H}_{cp}$ the correlation between them.
Since we aim at reasoning over pose-graphs, map points must be marginalized. The reduced Hessian matrix can be computed using the Schur complement as:
\begin{equation}
    \boldsymbol{H}_c' = \boldsymbol{H}_c -  \boldsymbol{H}_{cp} \ \boldsymbol{H}_p^{-1} \ \boldsymbol{H}_{cp}^T
\end{equation}

\begin{figure*}[t!]
    \centering
    \includegraphics[width=0.8\linewidth]{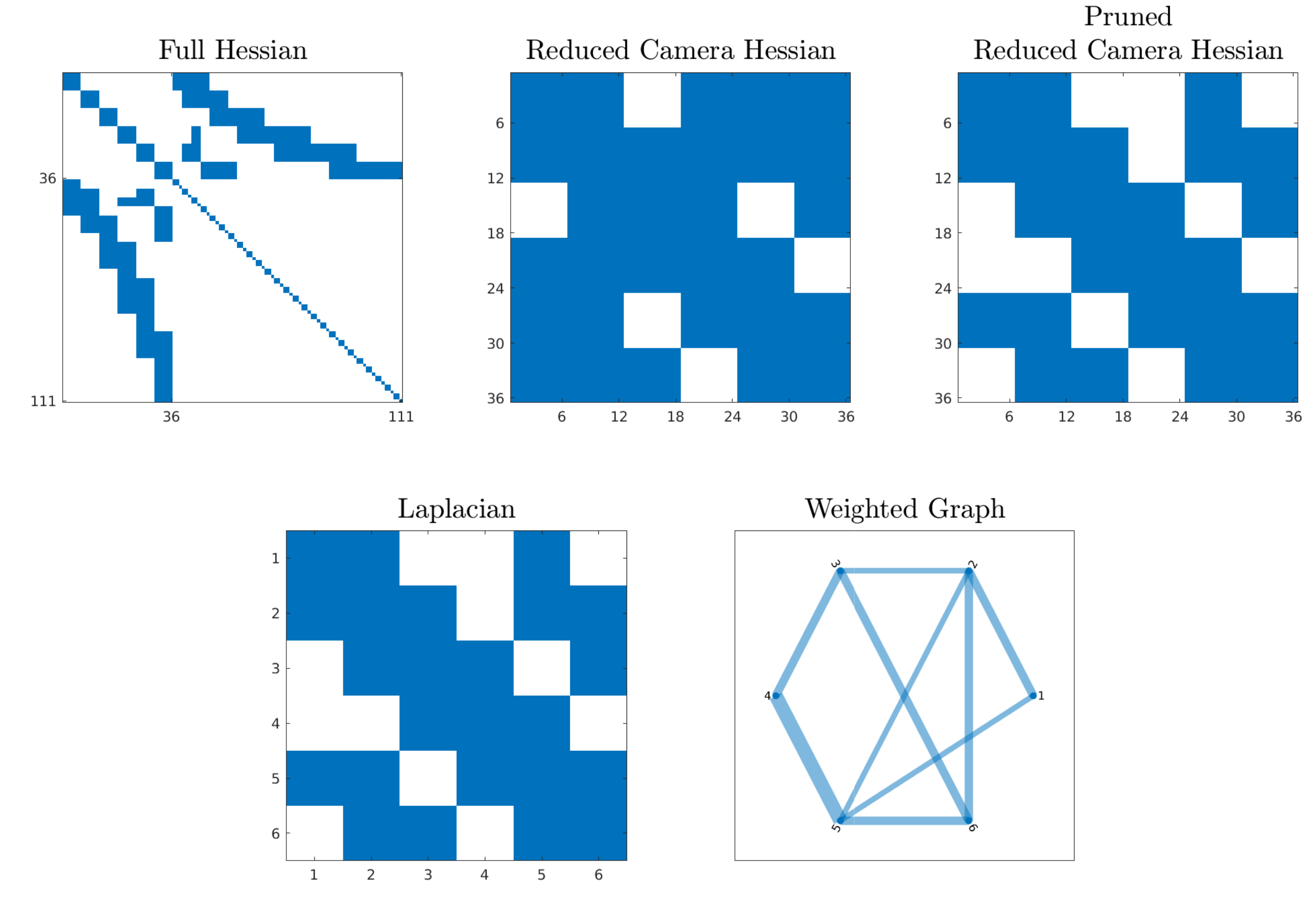}
    \caption{Sparsity patterns of the Hessian and Laplacian matrices in a toy example with 6 poses and 25 map points. From top left to bottom right: SLAM Hessian $\boldsymbol{H}_{\text{SLAM}}'$, reduced Hessian before and after pruning connections with less than 3 observations in common $\boldsymbol{H}_c'$, Laplacian and resulting weighted pose-graph.}
    \label{fig:hessian}
\end{figure*}

The block-sparsity pattern of the above matrix is sufficient to define the pose-graph topology.
The edge weights (dependent of the Hessian of the relative transformations) will be given by off-diagonal terms of $\boldsymbol{H}_c'$. To sparsify the pose-graph, we only keep those edges whose connected vertices share a minimum number of observations as in the essential graph, i.e., the intersection of the sets of observed landmarks is above a threshold. Figure~\ref{fig:hessian} contains an example of the sparsity patterns of the Hessians and the Laplacian, and the resulting graph.

\subsection{Frontier Detection}\label{SS:frontiers}

From the sparse 3D point cloud generated by ORB-SLAM, we build an Octomap and project it to the ground creating a 2D grid map in which frontiers can be spotted after several morphology transformations. Two frontiers detectors are used over the occupancy map, somehow following the popular work by Umari and Mukhopadhyay~\cite{umari17}. The first of them is based of rapidly-exploring random trees, while the second one employs the Canny edge detection algorithm. The candidates are clustered using the mean-shift algorithm, and then filtered to prune old points, low-informative returns and unreachable locations.
The main limitation of this module is that the candidate search is restricted to the ground surface, limiting the algorithm application to wheeled robots.
In future work, we plan to extend this module to detect 3D goals from sensor measurements, leveraging the insight that frontiers are bound to appear in recently sensed areas~\cite{keidar14}.

\subsection{Weighted Graph Hallucination}

\begin{figure}[t!]
    \centering
    \includegraphics[width=0.8\linewidth]{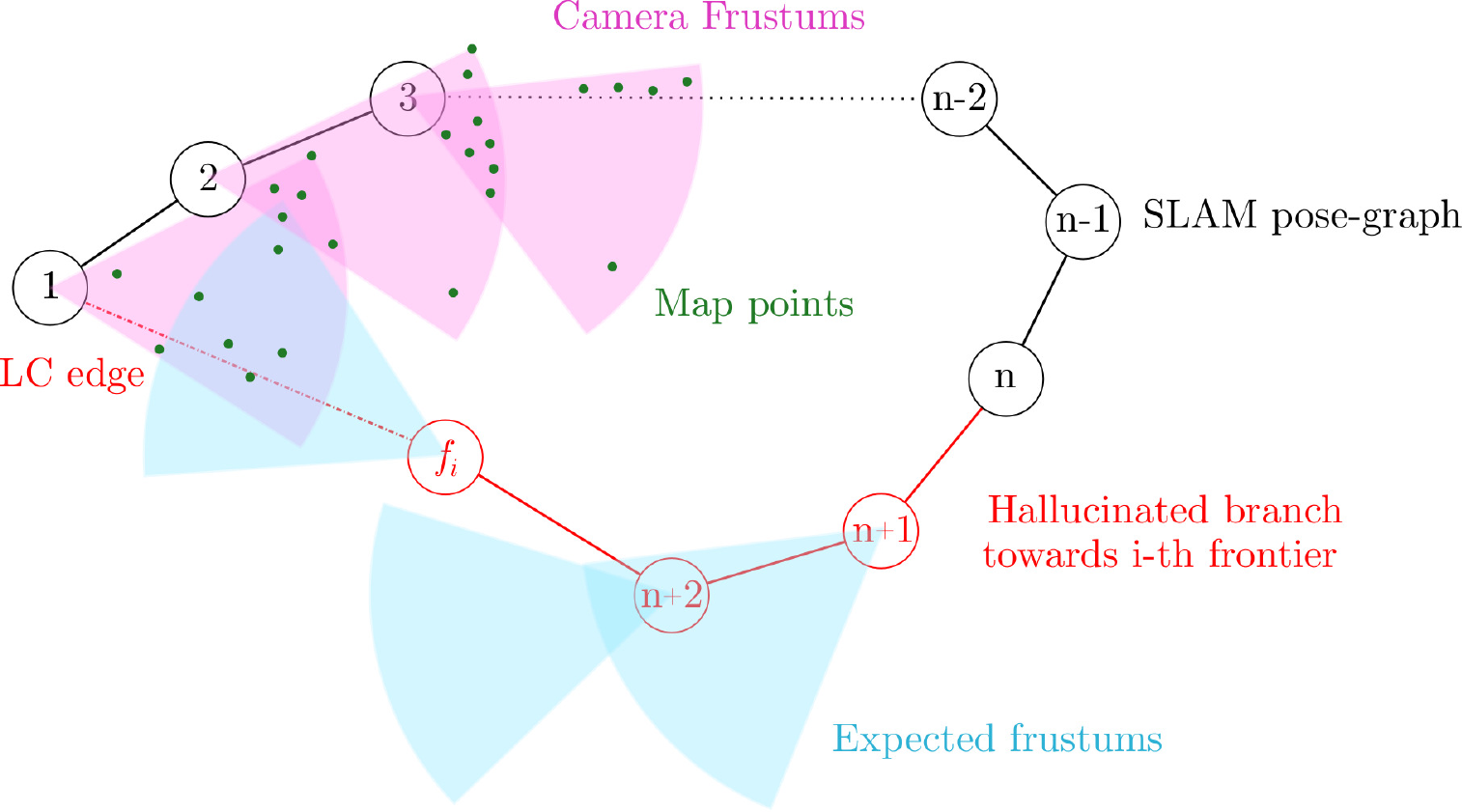}
    \caption{Example of the graph hallucination process towards frontier $f_i$, considering $n_{p,min}=3$ and $n_{p,max}=6$. A loop closure edge with probability $p^{LC}=1$ has been created between vertices ``$f_i$" and ``$1$".}
    \label{fig:graph_pred}
\end{figure}

\subsubsection{Graph prediction.} The pose-graph from ORB-SLAM is hallucinated towards every frontier spotted, thus creating as many graphs as frontiers. For each of them, we add a number of vertices along the expected path to reach the associated frontier; which is computed using Dijkstra's global planner from ROS navigation stack. The more complex or longer the path, the greater number of vertices will be in the hallucinated branch. We are aware that this method implies evaluating one single hypothesis for each frontier, although it has not been considered a critical issue. Initially, every vertex in the branch is just connected sequentially with its predecessor with an odometry constraint. After that, we consider the possibility of loop closures to appear. To do so, we first identify the set of existing map points expected to be observed from every hallucinated node (i.e., lie inside its expected frustum). If the number of covisible points with any other existing node in the SLAM graph, $n_p$, is higher than a certain threshold, $n_{p,min}$, the two will be connected by a loop closure edge. This edge will have the following probability of occurrence associated with it:
\begin{equation}
    p^{LC}=
    \begin{cases}
        0 \quad & \text{if} \quad n_p < n_{p,min} \\
        1 \quad & \text{if} \quad n_p > n_{p,max} \\
        \frac{n_p}{n_{p,max}}\quad & \text{otherwise} \\
    \end{cases} \label{eq:lc_prob}
\end{equation}
where $n_{p,max}$ is a defined upper bound. Figure~\ref{fig:graph_pred} illustrates the process of graph prediction with a toy example.

\subsubsection{Hessian estimation.} Once the topology of the hallucinated graph is defined, the FIMs associated to each edge in the hallucinated branch must be computed in order to later weight the graph. Assuming that visual odometry uncertainty will remain similar unless a loop closing occurs, the relative Hessians of odometry edges can be considered equal to the last one from the known SLAM pose-graph:
\begin{equation}
    \boldsymbol{H}^{odom} = \boldsymbol{H}^{prev} \label{eq:Hodom}
\end{equation}

In the case that a loop closure is expected to occur between a pair of vertices, the Hessian will be given instead by the Jacobian matrices of the reprojection error of all covisible points and the likelihood of closing the loop, as shown in~\eqref{eq:lc_prob}:
\begin{equation}
    \boldsymbol{H}^{LC} = p^{LC} \sum_i^{n_p} \boldsymbol{J}_i^T\boldsymbol{J}_i \label{eq:Hlc}
\end{equation}
Hence, far and scarce covisible points will result in higher expected uncertainties.

The previous Hessians, however, do not account for the decrease in the environment's uncertainty as new regions are explored, since the complete set of landmarks is unknown. To also include it and motivate a balance between exploration and exploitation, we leverage the insight of discounting information from Carrillo \textit{et al.}~\cite{carrillo18}. We adapt this concept to operate in the task space (i.e., uncertainty) rather than in the information space (i.e., entropy).
Besides, instead of discounting from the map's entropy the information value of visiting a region with high localization uncertainty as in~\cite{carrillo18}; our approach builds upon the idea of penalizing the above Hessians if no new areas are visited. Just like the Shannon-R\'enyi entropy, the Hessian of the $j$-th edge will be given by:
\begin{equation}
    \boldsymbol{H}_j = \boldsymbol{H} - \left(\frac{1}{1-\alpha}\right) \boldsymbol{H} \label{eq:Hedge}
\end{equation}
\noindent where $\boldsymbol{H}$ is to be computed using~\eqref{eq:Hodom} or~\eqref{eq:Hlc} depending on the constraint type,  $\alpha=1+\frac{1}{\sigma}$ and $\sigma$ is a parameter that encodes the novelty of the regions to visit. More specifically, it is computed as the percentage of unseen area expected to be observed in the occupancy grid map within a $1.5$ meter radius around the node.

\subsubsection{Graph weighting.} Finally, every edge in the hallucinated pose-graph is weighted with the \textit{D-opt} of the corresponding Hessian, i.e., $\gamma_j = \|\boldsymbol{H}_j\|_0$; so as to allow the use of~\eqref{eq:dopt_graph}.

\subsection{Frontier Selection and Navigation}

In order to identify the optimal frontier, we compute \textit{D-opt} for each hallucinated graph using~\eqref{eq:dopt_graph} and select that with the highest value:
\begin{equation}
    f^* = \argmax_f D\text{-}opt(\boldsymbol{L}_\gamma(f))
\end{equation}
with $\boldsymbol{L}_\gamma(f)$ the weighted Laplacian of the hallucinated graph to a given frontier.

The above optimization (which entails the graph hallucination and Hessian estimation processes) implicitly penalizes visiting distant candidates and encourages a balance between the uncertainty decrease in both the robot location and the map that occurs when reobserving known landmarks, and the increase of knowledge about the environment when visiting new regions.

Since the global plan was already computed, navigation comes down to following that path, a task inherent of the local planner. We use the time elastic band approach~\cite{rosmann17}, which optimizes the trajectory locally with respect to different constraints. As occurred with the global planner, the local planner acts over a cost map built from lidar measurements.

\subsection{Recovery Behavior}

As shown in Figure~\ref{fig:active_system}, navigation goals can be exceptionally obtained from the recovery module if ORB-SLAM gets lost during exploration. In such cases, the robot uses the wheel odometry to localize itself and generates navigation goals to previously-visited areas in order to facilitate relocalization. Since most usual tracking losses are due to getting excessively close to an obstacle, the first goal consists of a $180\degree$ rotation. If no relocalization occurs, the robot is commanded to previously-visited locations with high relocalization potential. To identify them, we search all the pose-graph nodes within a $2~m$ radius and compute the number of map points visible from each of them. The preferred destination will be that with highest map point density. 

\begin{figure}[t!]
    \centering
    \includegraphics[width=0.8\linewidth]{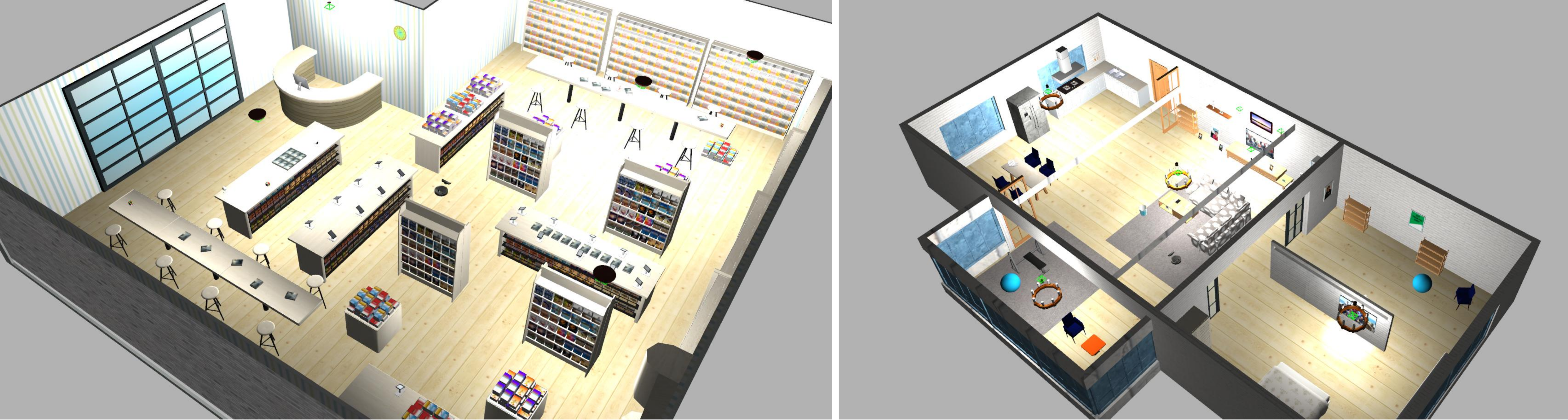}
    \caption{View in Gazebo of the AWS bookstore (left) and house (right) scenarios.}
    \label{fig:envs}
\end{figure}

\section{Experimental Validation}\label{S:4}

The system has been tested in Gazebo simulator. The experiment consists of autonomously exploring AWS bookstore and house scenarios\footnote{https://github.com/aws-robotics/}, which texture is rich enough to be processed by visual SLAM algorithms; see Figure \ref{fig:envs}. Termination condition is the absence of candidate frontiers. We used a wheeled robot with a Kinect RGB-D camera (for SLAM) and a lidar (for path planning and safe navigation).
All experiments have been performed in Ubuntu 20.04 and ROS Noetic, using an Intel Core i9-10900K CPU and a NVidia GeForce RTX 3070 GPU.
For more details on the configuration parameters of each module, we refer the reader to the \href{https://github.com/JulioPlaced/ExplORB-SLAM
}{project repository}.

\begin{figure}[t!]
    \centering
    \includegraphics[max width=0.4\linewidth, max height=3.5cm]{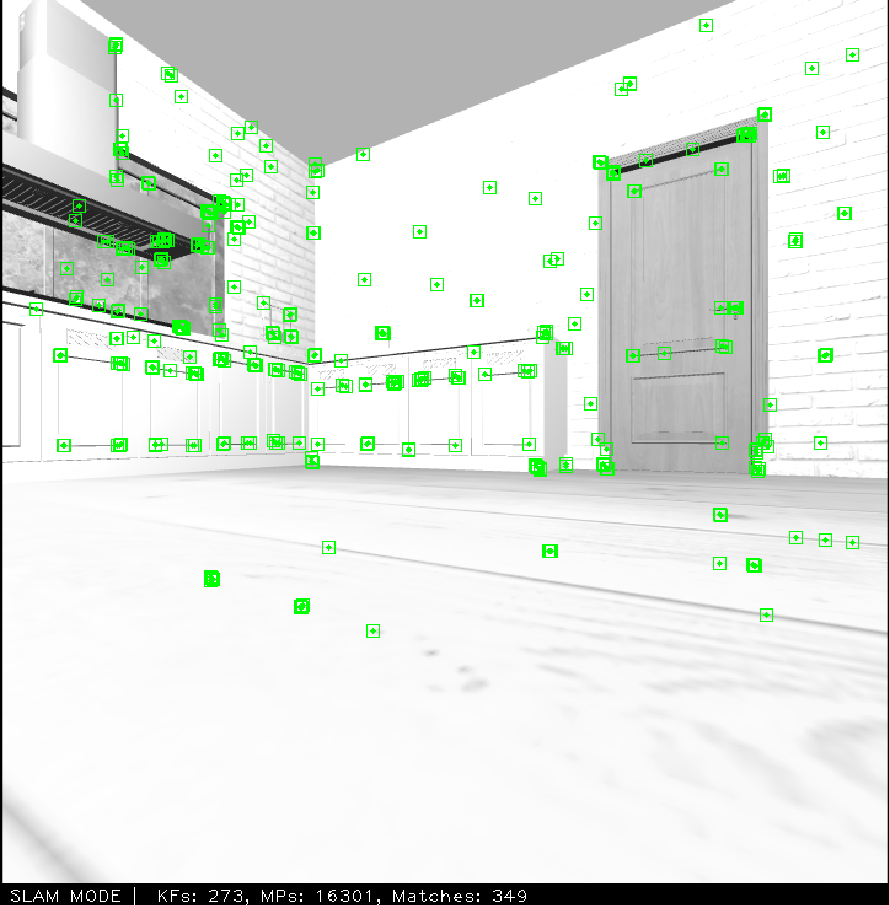} \hfill
    \includegraphics[max width=0.6\linewidth, max height=3.5cm]{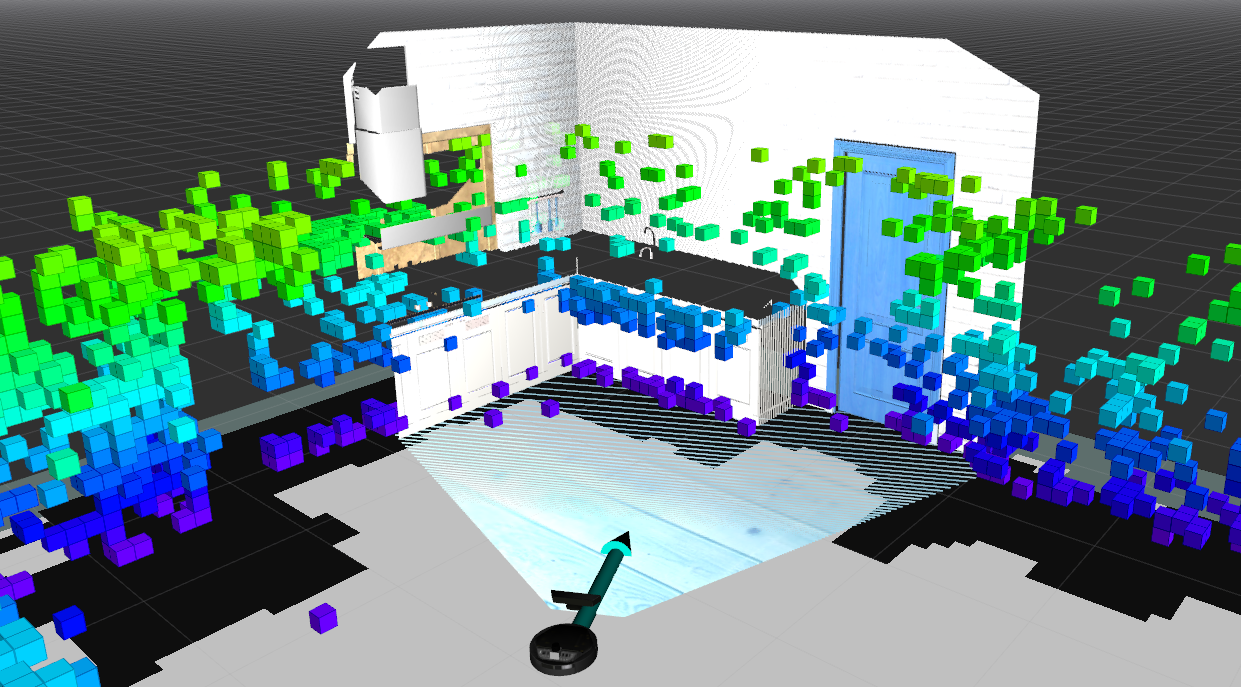}
    \caption{SLAM image with matched landmarks, Octomap and grid map built.}
    \label{fig:gridmap_examples}
\end{figure}
\begin{figure}[t!]
    \centering
    \includegraphics[max width=0.32\linewidth, max height=5cm]{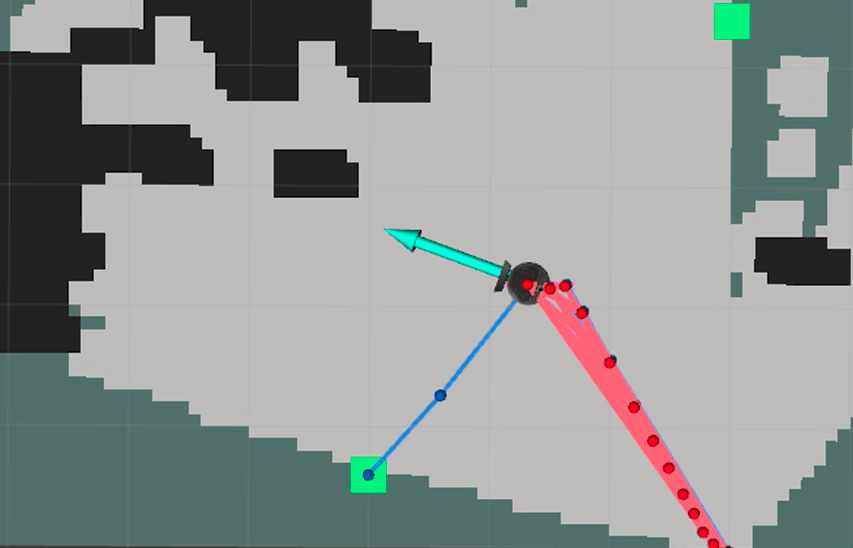} \hfill
    \includegraphics[max width=0.32\linewidth, max height=5cm]{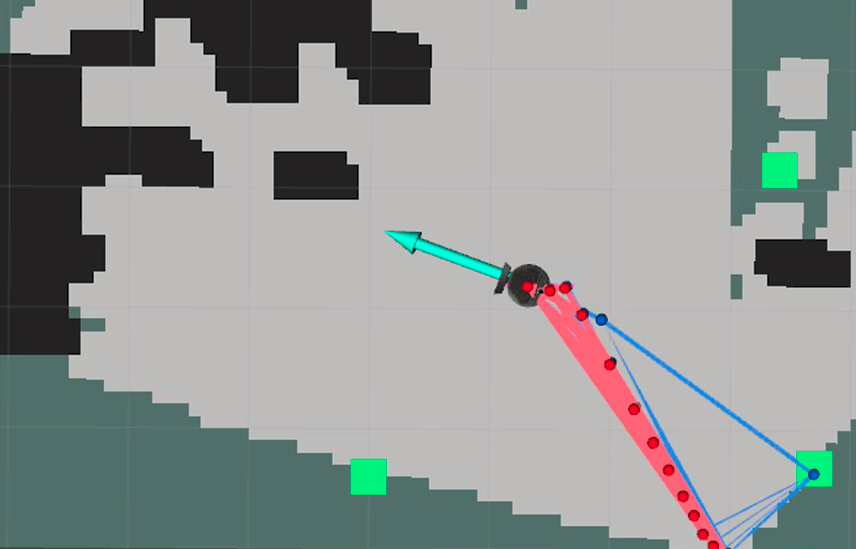} \hfill
    \includegraphics[max width=0.32\linewidth, max height=5cm]{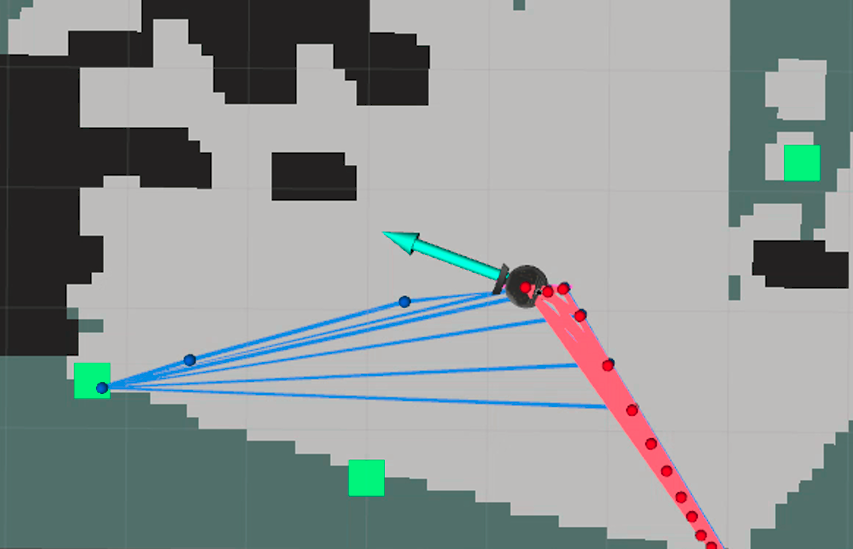} \hfill
    \caption{Examples of the graph hallucination process during exploration.}
    \label{fig:hallucination_examples}
\end{figure}

To better show the proposed system, figures~\ref{fig:gridmap_examples} and~\ref{fig:hallucination_examples} contain two operating examples of the different modules during exploration of the house environment.
In the former, we present the image fed to ORB-SLAM and the landmarks detected (left). Right image contains the Octomap built from those landmarks, on top of the depth image point cloud. For visualization purposes, Octomap height has been restricted between $0.1$ and $2 m$. Also, we include the rectified grid map and the estimated robot location (blue arrow).
On the other hand, Figure~\ref{fig:hallucination_examples} illustrates the graph hallucination process towards three different frontiers (green squares). In all cases, the SLAM (red) and hallucinated (blue) pose-graphs have been plotted on top of the rectified grid map. While the first frontier does not produce reobservation edges in the hallucinated graph, the second and third do; showing that along the path to reach them, a greater number of known landmarks is expected to be observed.

\begin{figure}[t!]
    \centering
    \includegraphics[max width=0.48\linewidth, max height=3.1cm]{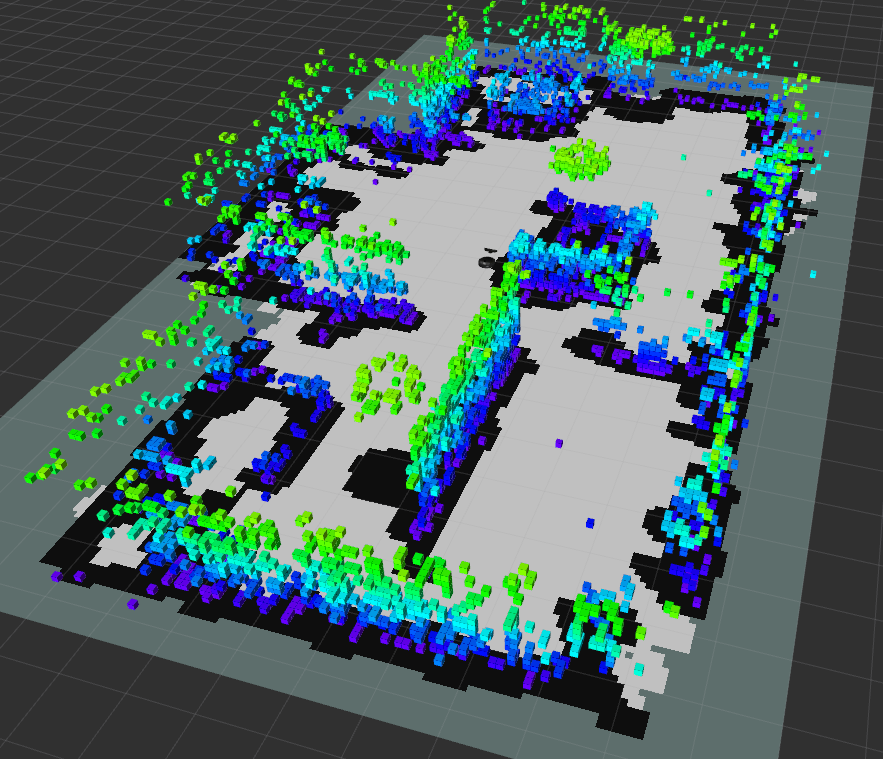} \qquad
    \includegraphics[max width=0.48\linewidth, max height=3.1cm]{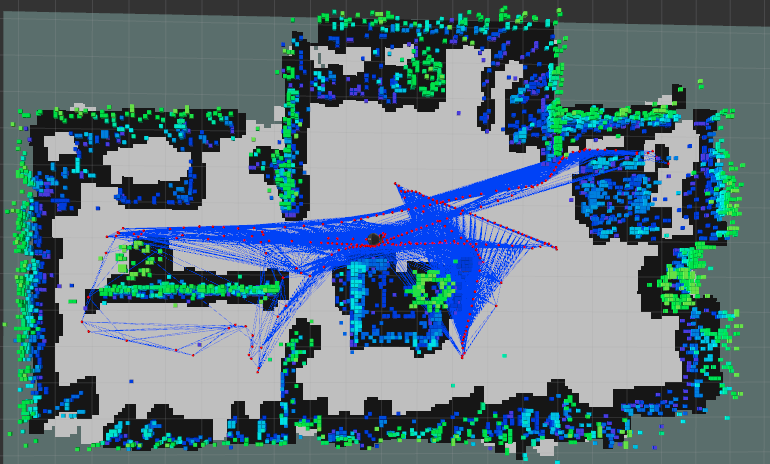}
    \caption{Maps and pose-graphs generated during exploration of house scenario.}
    \label{fig:results_maps1}
\end{figure}

\begin{figure}[t!]
    \centering
    \includegraphics[max width=0.48\linewidth, max height=3.1cm]{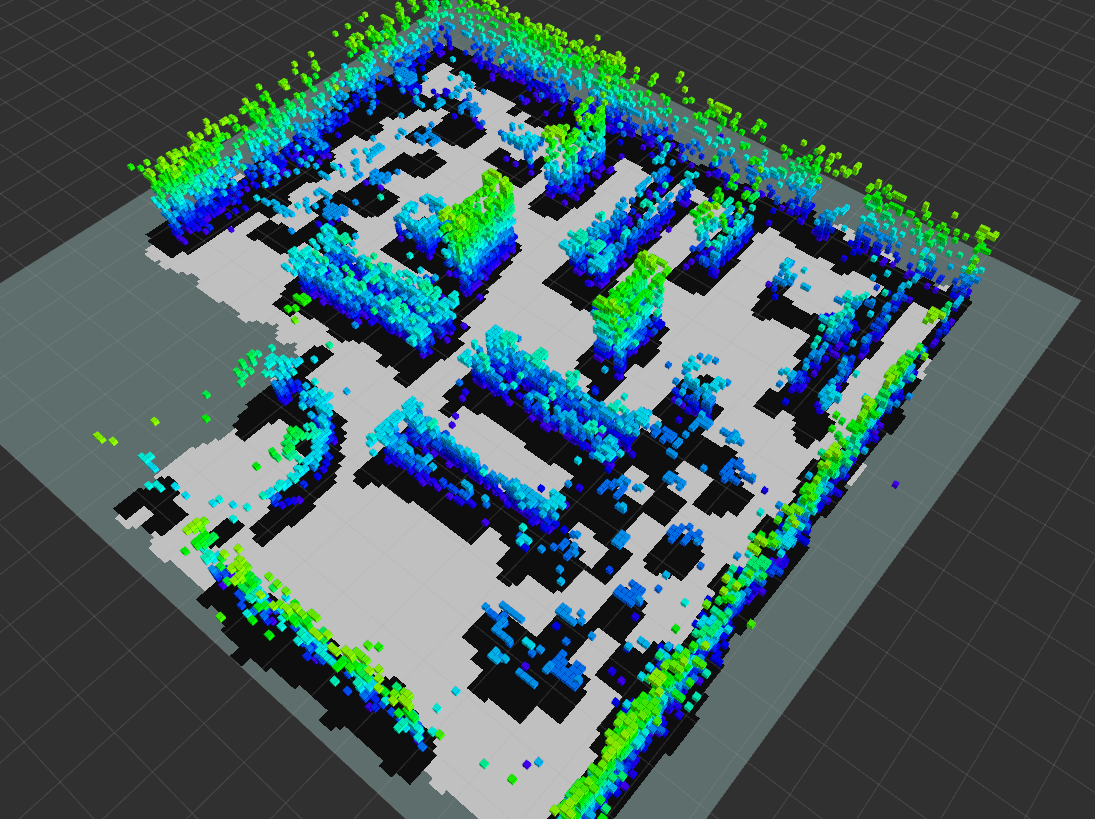} \qquad
    \includegraphics[max width=0.48\linewidth, max height=3.1cm]{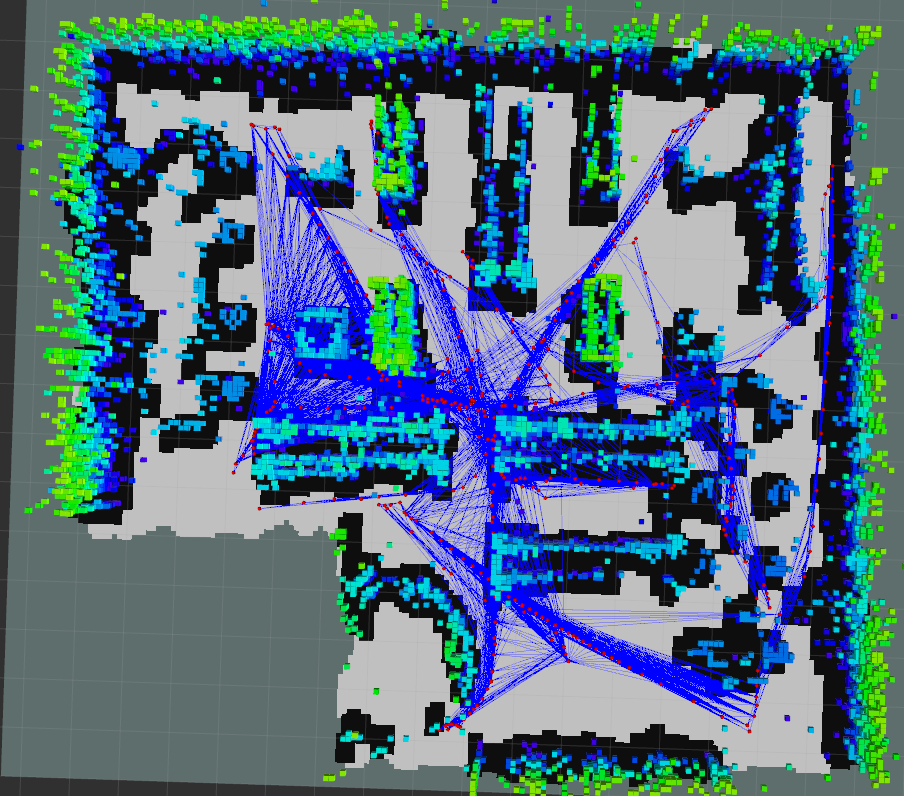}
    \caption{Maps and pose-graphs generated during exploration of bookstore scenario.}
    \label{fig:results_maps2}
\end{figure}

Finally, figures~\ref{fig:results_maps1} and~\ref{fig:results_maps2} show the resulting maps and pose-graphs after the exploration of the two environments. In the first one, exploration lasted $560 s$ and ORB-SLAM was able to close two loops ---thus performing two graph optimizations.
In the second one, the time increased to $984 s$ and 5 optimizations were performed. Decision making represented only $5\%$ and $7.3\%$ of these times, respectively. Both environments were fully explored, despite the presence of open regions in the grid maps that correspond to low-textured walls in the simulator. Since planning feasibility is checked for all candidates, no frontiers were detected in these areas.

\section{Conclusions}\label{S:5}

In this work, we have presented an open-source active SLAM method based on ORB-SLAM2, a visual graph-SLAM system that creates a landmark representation of the environment and accurately estimates the robot trajectory.
We employ that information to perform decision-making over graphs, which involves: (i) identifying the candidate destinations, (ii) hallucinating the existing pose-graph towards those destinations and weighting their edges appropriately, and (iii) selecting the $D$-optimal candidate by analyzing the topology of the graphs. Experiments conducted on realistic simulated environments demonstrate the performance of the method. As future work, we plan to study multi-robot exploration behavior and conduct 3D experiments.

\section*{Acknowledgments}
This work was supported by the Spanish government under grant PID2019‐108398GB‐I00 and by Arag\'on government under grant T45-20R.

%
%


\begin{thebibliography}{26}
%

\bibitem {placed22}
Placed, J. A., Strader J., Carrillo H., Atanasov N., Indelman V., Carlone L., and Castellanos, J. A. (2022). A survey on active simultaneous localization and mapping: State of the art and new frontiers. arXiv preprint arXiv: 2207.00254.

\bibitem {makarenko02}
Makarenko, A. A., Williams, S. B., Bourgault, F., and Durrant-Whyte, H. F. (2002). An experiment in integrated exploration. In \textit{IEEE/RSJ International Conference on Intelligent Robots and Systems} (Vol. 1, pp. 534-539).

\bibitem{valencia12}
Valencia, R., Miró, J. V., Dissanayake, G., and Andrade-Cetto, J. (2012). Active pose SLAM. In \textit{IEEE/RSJ International Conference on Intelligent Robots and Systems} (pp. 1885-1891).

\bibitem {umari17}
Umari, H., and Mukhopadhyay, S. (2017). Autonomous robotic exploration based on multiple rapidly-exploring randomized trees. In \textit{IEEE/RSJ International Conference on Intelligent Robots and Systems} (pp. 1396-1402).

\bibitem{carrillo18}
Carrillo, H., Dames, P., Kumar, V., and Castellanos, J. A. (2018). Autonomous robotic exploration using a utility function based on Rényi’s general theory of entropy. \textit{Autonomous Robots}, 42(2), 235-256.

\bibitem {feder99}
Feder, H. J. S., Leonard, J. J., and Smith, C. M. (1999). Adaptive mobile robot navigation and mapping. \textit{The International Journal of Robotics Research}, 18(7), 650-668.


\bibitem{deng20}
Deng, D., Xu, Z., Zhao, W., and Shimada, K. (2020). Frontier-based automatic-differentiable information gain measure for robotic exploration of unknown 3D environments. arXiv preprint arXiv:2011.05288.

\bibitem {carlone17}
Carlone, L., and Karaman, S. (2018). Attention and anticipation in fast visual-inertial navigation. \textit{IEEE Transactions on Robotics}, 35(1), 1-20.

\bibitem{wang20}
Wang, J., and Englot, B. (2020). Autonomous exploration with expectation-maximization. \textit{Robotics Research} (pp. 759-774). Springer.

\bibitem{placed21iros}
Placed, J. A., and Castellanos, J. A. (2021). Fast Autonomous Robotic Exploration Using the Underlying Graph Structure. In \textit{IEEE/RSJ International Conference on Intelligent Robots and Systems} (pp. 6672-6679).

\bibitem{khosoussi14}
Khosoussi, K., Huang, S., and Dissanayake, G. (2014). Novel insights into the impact of graph structure on SLAM. In \textit{IEEE/RSJ International Conference on Intelligent Robots and Systems} (pp. 2707-2714).

\bibitem{khosoussi19}
Khosoussi, K., Giamou, M., Sukhatme, G. S., Huang, S., Dissanayake, G., and How, J. P. (2019). Reliable graphs for SLAM. \textit{The International Journal of Robotics Research}, 38(2-3), 260-298.

\bibitem{placed21}
Placed, J. A., and Castellanos, J. A. (2021). A General Relationship between Optimality Criteria and Connectivity Indices for Active Graph-SLAM . arXiv preprint arXiv:2110.01289.

\bibitem{chen20}
Chen, Y., Huang, S., and Fitch, R. (2020). Active SLAM for mobile robots with area coverage and obstacle avoidance. \textit{IEEE/ASME Transactions on Mechatronics}, 25(3), 1182-1192.

\bibitem{mur17}
Mur-Artal, R., and Tardós, J. D. (2017). ORB-SLAM2: An open-source SLAM system for monocular, stereo, and RGB-D cameras. \textit{IEEE Transactions on Robotics}, 33(5), 1255-1262.

\bibitem{kiefer74}
Kiefer, J. (1974). General equivalence theory for optimum designs (approximate theory). \textit{The annals of Statistics}, 849-879.

\bibitem{triggs99}
Triggs, B., McLauchlan, P. F., Hartley, R. I., and Fitzgibbon, A. W. (1999, September). Bundle adjustment—a modern synthesis. In \textit{International Workshop on Vision Algorithms} (pp. 298-372). Springer, Berlin, Heidelberg.

\bibitem{keidar14}
Keidar, M., and Kaminka, G. A. (2014). Efficient frontier detection for robot exploration. \textit{The International Journal of Robotics Research}, 33(2), 215-236.

\bibitem{rosmann17}
Rösmann, C., Hoffmann, F., and Bertram, T. (2017). Integrated online trajectory planning and optimization in distinctive topologies. \textit{Robotics and Autonomous Systems}, 88, 142-153.

\end{thebibliography}
\end{document}